\documentclass{article}
\usepackage{amsmath,epsfig}
\usepackage[preprint]{spconfa4}
\usepackage{amssymb}
\usepackage{booktabs}
\usepackage{multirow}
\usepackage{setspace}
\usepackage{subfigure}

\usepackage{color}

\copyrightnotice{978-1-6654-3864-3/21/\$31.00~\copyright 2021 IEEE}

\let\OLDthebibliography\thebibliography
\renewcommand\thebibliography[1]{
  \OLDthebibliography{#1}
  \setlength{\parskip}{0pt}
  \setlength{\itemsep}{0pt plus 0.3ex}
}

\begin{document}
 \topmargin=0mm

\def\x{{\mathbf x}}
\def\L{{\cal L}}

\title{Hands-on Guidance for Distilling Object Detectors}
%
\name{
Yangyang Qin, Hefei Ling$^{\dagger}$\thanks{$^{\dagger}$Corresponding author. This work was supported in part by the Natural Science Foundation of China under Grant 61972169 and U1536203, in part by the National key research and development program of China(2016QY01W0200), in part by the Major Scientific and Technological Project of Hubei Province (2018AAA068 and 2019AAA051).},
Zhenghai He, Yuxuan Shi, Lei Wu}

\address{Department of Computer Science and Technology, Huazhong University of Science and Technology}

\maketitle

\begin{abstract}
Knowledge distillation can lead to deploy-friendly networks against the plagued computational complexity problem, but previous methods neglect the feature hierarchy in detectors.
Motivated by this, we propose a general framework for detection distillation. Our method, called \emph{Hands-on Guidance Distillation}, distills the latent knowledge of all stage features for imposing more comprehensive supervision, and focuses on the essence simultaneously for promoting more intense knowledge absorption. Specifically, a series of novel mechanisms are designed elaborately, including correspondence establishment for consistency,  hands-on imitation loss measure and re-weighted optimization from both micro and macro perspectives.
We conduct extensive evaluations with different distillation configurations over VOC and COCO datasets, which show better performance on accuracy and speed trade-offs.
Meanwhile, feasibility experiments on different structural networks further prove the robustness of our HGD.

\end{abstract}

\section{Introduction}
\label{sec:introduction}

Object detection, as a fundamental task of computer vision, of which deployment demand is growing steadily with deep learning development. However, the latest CNN-based detectors always gain extreme precision at the cost of massive memory and computational budget.
It makes the deployment arduous on resource-constrained devices such as mobile phones. Consequently, the trade-off between accuracy and overhead has become a hot issue, and some model compression methods~\cite{han2015learning,han2015deep,romero2014fitnets,hinton2015distilling,rastegari2016xnor,wu2016quantized} have been proposed, including network pruning, knowledge distillation, and quantization.

Unlike other compression methods that require demanding hardware or software environments to get inference acceleration, knowledge distillation is a simple method to learn a thin but high-performance student model under the guidance of the teacher network.
However, the initial works~\cite{romero2014fitnets,hinton2015distilling} are not designed for the object detection task but the classification task. As a basic task, detection involves fewer classes, but additional analysis of accurate location is required, resulting in the little effect of distilling inter-class similarity knowledge through softened outputs~\cite{wang2019distilling}.  
As widely summarized, the best detector must rely on the best feature extractor to predict the objectness scores and the locations. Therefore, more and more researchers~\cite{wang2019distilling,chen2017learning,zhu2019mask} turn their attention to the feature imitation for detection knowledge distillation.

Overall, current feature imitation methods all focus on capturing the knowledge from the high-level features near the detection heads, from which the thin model tries to learn similar predictions.
Obviously, these parts have the highest level semantics from a series of convolutions and directly impact the final detection part.
But in our opinion, this kind of imitation mode is not comprehensive enough.
Only learning the highest-level semantics, student models fail to fully view what to learn and even overfit sometimes. To mitigate this gap, we propose a new \emph{Hands-on Guidance Distillation} (HGD) approach to fully utilize the teacher's knowledge.

\begin{figure}
    \centering
    \subfigure[Traditional guidance]{
        \begin{minipage}[t]{0.9\linewidth}
            \centering
            \includegraphics[width=0.9\linewidth]{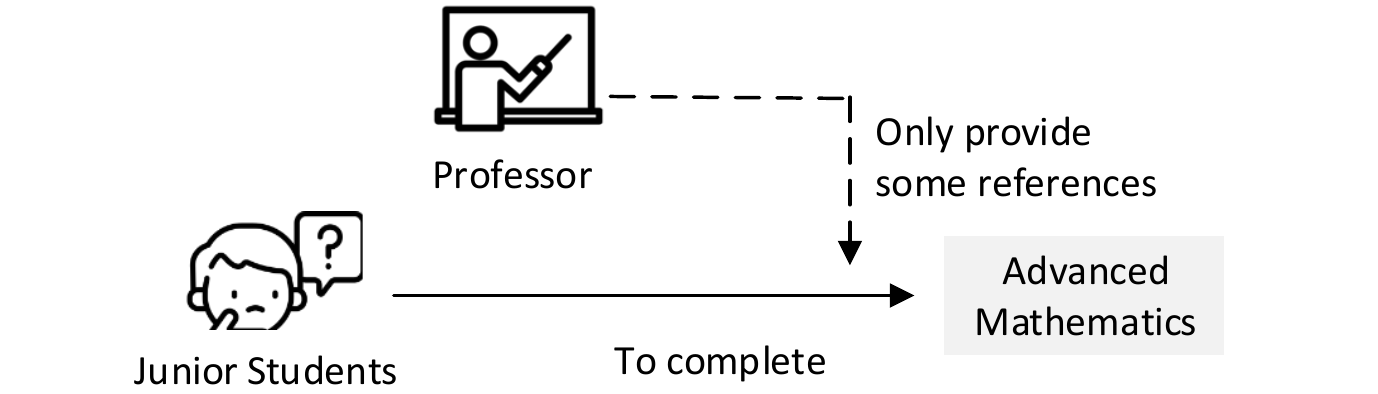}
        \end{minipage}%
    }\vspace{-1.8mm}
    ~
    \subfigure[Hands-on guidance]{
        \begin{minipage}[t]{0.9\linewidth}
            \centering
            \includegraphics[width=0.9\linewidth]{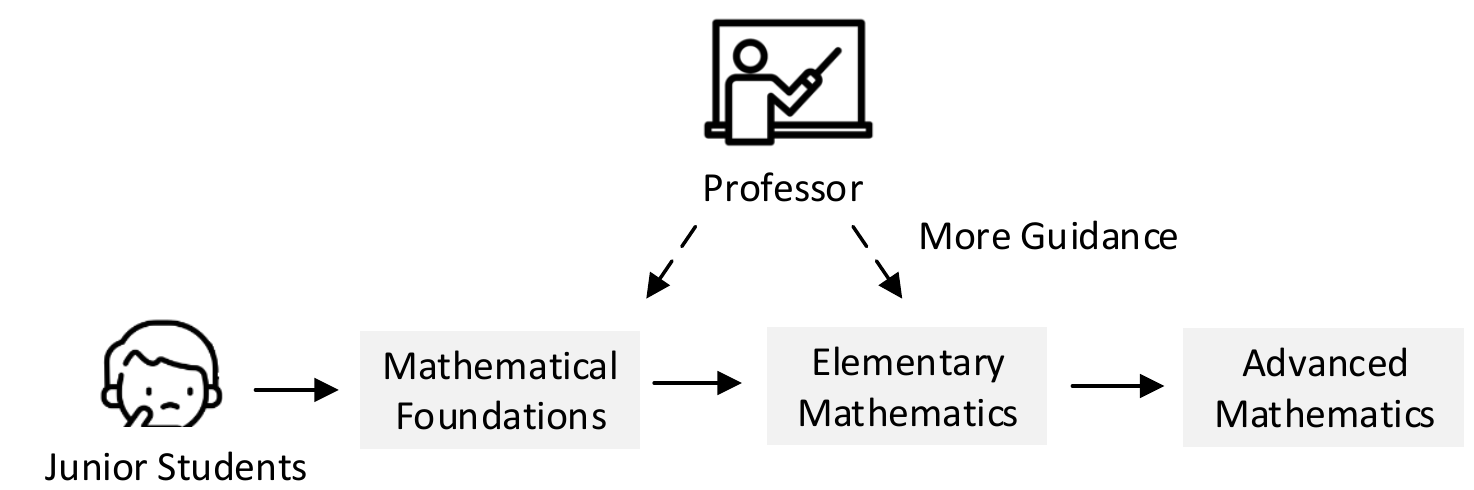}
        \end{minipage}
        \label{fig:HG_sample}
    }
    \vspace{-2mm}
    \caption{A vivid illustration on the principle of our method.
    With the detailed guidance  of the teacher step by step, the student can better solve challenging  problems. }
    \label{fig:sample}
    \vspace{-4mm}

\end{figure}

Before going into our approach's details, let's see a simple teaching example in Figure~\textcolor{red}{\ref{fig:sample}}.
Model training is like solving a problem. In the traditional distillation approach, student models are not told what to learn until they are close to the training end. Therefore, it's only slightly better than provided with ground truths (i.e. dataset supervision) in the end, and tends to fail when dealing with subtle tasks like object detection. It inspires us to reflect, why not give more guidance? That is, teaching the student model step by step during the training procedure. Benefit from it, thin student models may be able to solve challenging  problems without much effort.

Back to object detector, although the semantics of the high-level feature is rich to distinguish objects, we argue that more fine-grained location information implied by early-stage parts is equally important, as same idea also seen in the feature pyramid network~\cite{lin2017feature}. Furthermore, the early-stage features are the basis of later calculations so that the student can learn more about the work route. Therefore, our approach mainly distills each stage of the teacher model into student models as much as possible, which aims to comprehend the essence and accelerate convergence as shown in later experiments.
Considering there are too many features at all stages, we explore some ingenious strategies to concentrate on the emphasis, and verify them with experiments.
In summary, we make three main contributions as follows:
\vspace{-2 mm}
\renewcommand{\baselinestretch}{1.0}{
\begin{itemize}
\setlength{\itemsep}{0pt}
\setlength{\parsep}{0pt}
\setlength{\parskip}{0pt}
\item We propose a new distillation meta-architecture for the challenging detection task, and design to distill the exhaustive and refined prior knowledge from the teacher.
    It has more versatility for relying on only feature maps instead of anchors, labels, RPN, etc.
\item We explore some meticulous strategies more in line with our framework to achieve optimum, including feature matching strategy, improved loss function, and re-weighted optimization, complementing each other.
\item We conduct extensive comparison experiments, including 7.0\% more improvement of SSD-half on PASCAL VOC, and 1.9\% improvement on more complicated MS COCO datasets, which show better results than other methods.
    More experiments can be seen in Section~\ref{sec:experiment}.

\end{itemize}
}


\section{Related Work}
\label{sec:related_work}

\textbf{Lightweight Object Detection.}
 Since the first successful introduction of CNN~\cite{krizhevsky2017imagenet}, a variety of CNN-based detectors have been proposed in object detection, such as Faster R-CNN~\cite{ren2015faster} and SSD~\cite{liu2016ssd},  RetinaNet~\cite{lin2017focal}, which can be simply divided into two-stage and one-stage detectors.
 %
 Though the one-stage detectors can do real-time inference on GPU with competitive accuracy, considerable computational overhead is still needed. Thus, some researchers propose lightweight detectors~\cite{howard2017mobilenets,zhang2018shufflenet} with faster speed and more efficient structure for mobile devices.
  To further accelerate CNN models, a large number of model compression methods come into being. For example, network pruning~\cite{han2015learning,han2015deep,park2016faster} removes the unnecessary computation redundancy, and quantization~\cite{rastegari2016xnor,wu2016quantized} relies on low-precision computation for acceleration. 
  Different from the methods above that require many restrictions, knowledge distillation is a more straightforward  method.
 \\
\textbf{Knowledge Distillation.}
 Knowledge distillation is orthogonal to other model compression methods.
 It aims to increase the precision of deploy-friendly networks with the help of the high precision network.
 Hinton first introduced the seminal distillation work through soft output learning~\cite{hinton2015distilling}, while hint learning~\cite{romero2014fitnets} explored from another perspective of learning intermediate representations as hints. Soon afterward, various approaches were proposed to dig out the potential hints, such as attention map~\cite{zagoruyko2016paying}, internal relationship~\cite{yim2017gift,park2019relational}, dataset~\cite{chen2019data,wang2018dataset}, etc.
 And~\cite{cheng2020explaining} used task-irrelevant visual concepts to explain why distillation works.
 Recently, some researchers turn their attention from the classification to the challenging detection distillation research. \cite{chen2017learning,li2017mimicking} combined both high-level feature imitation and final output distillation, and \cite{wang2019distilling,zhu2019mask} tried to imitate the central part of the feature map instead of the entire content.
 Unlike works  mentioned above, we explore a novel distillation architecture to make full use of all teacher model features rather than only high-level features to help student models comprehend the essence easily.


\begin{figure}[]
    \setlength{\abovecaptionskip}{5pt}
    \centering
    \vspace{-3mm}
    \subfigure[\footnotesize Original Distillation\cite{hinton2015distilling}]{
        \label{fig:KD_sample}
        \begin{minipage}[t]{0.43\linewidth}
            \centering
            \includegraphics[width=0.9\linewidth]{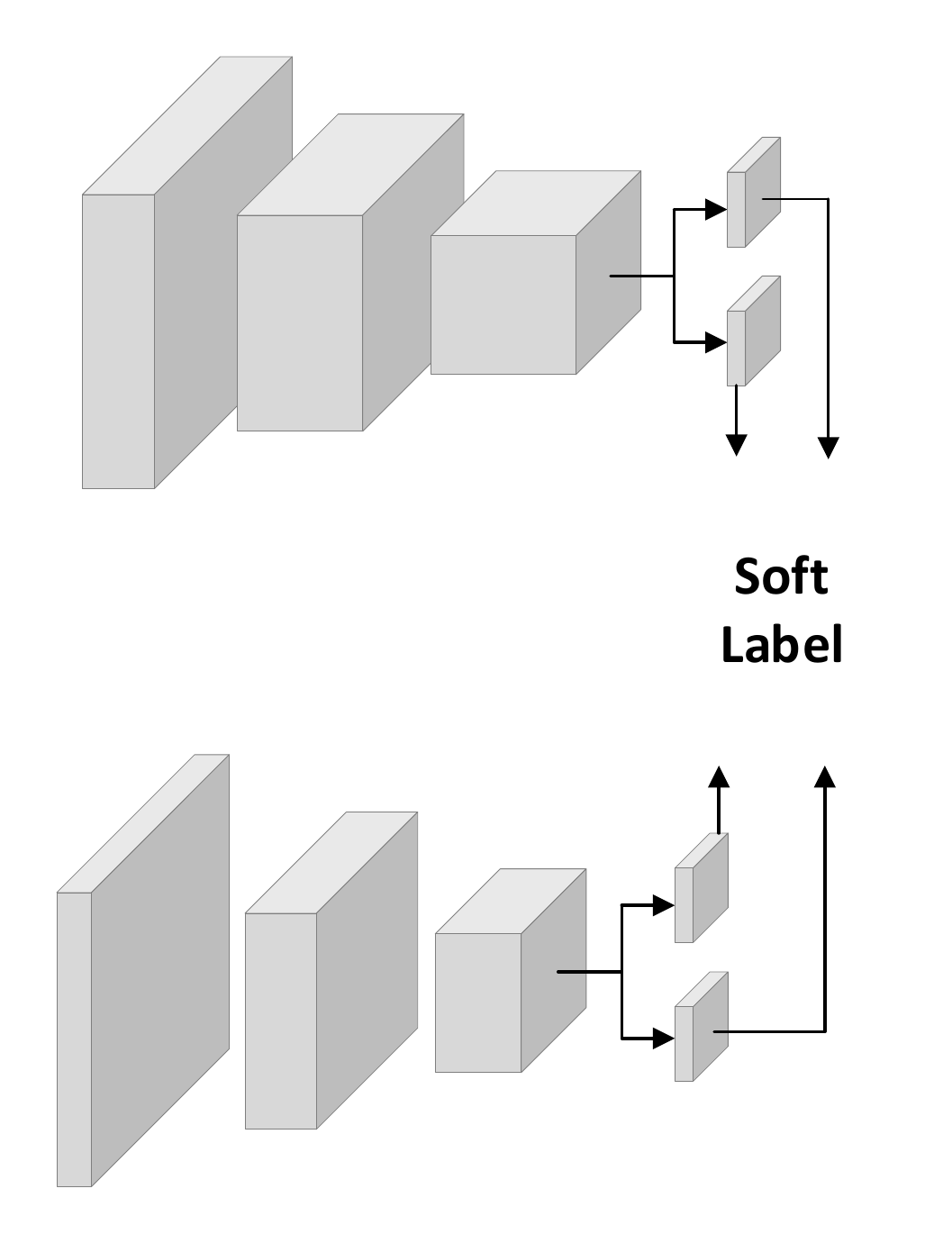}
        \end{minipage}%
    }
    ~
    \subfigure[\footnotesize Detection Distillation\cite{chen2017learning}]{
        \begin{minipage}[t]{0.43\linewidth}
            \centering
            \includegraphics[width=0.9\linewidth]{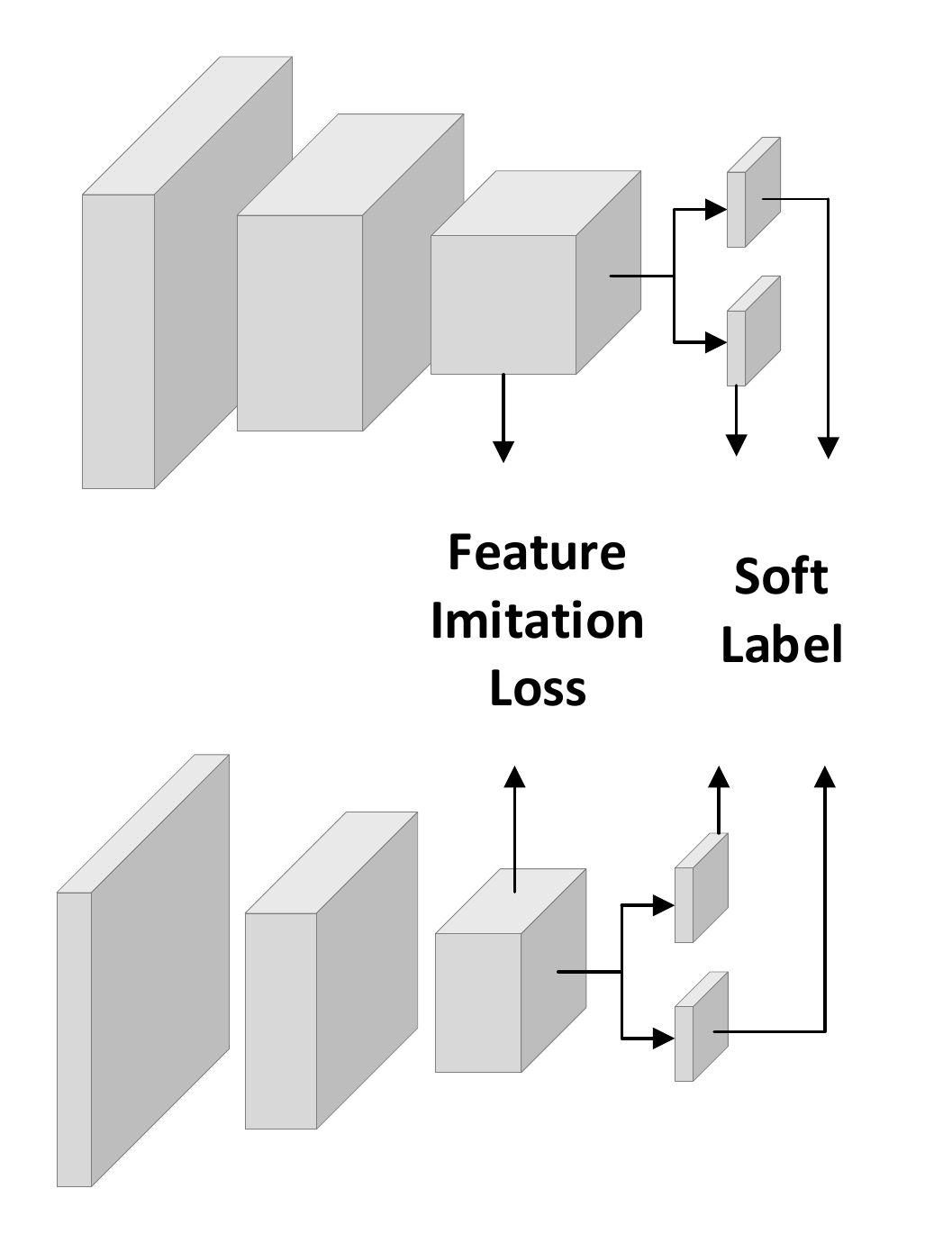}
        \end{minipage}
    }\vspace{-1mm}
    ~
    \subfigure[\footnotesize Fine-grained Distillation\cite{wang2019distilling,zhu2019mask}]{
        \begin{minipage}[t]{0.43\linewidth}
            \centering
            \includegraphics[width=0.9\linewidth]{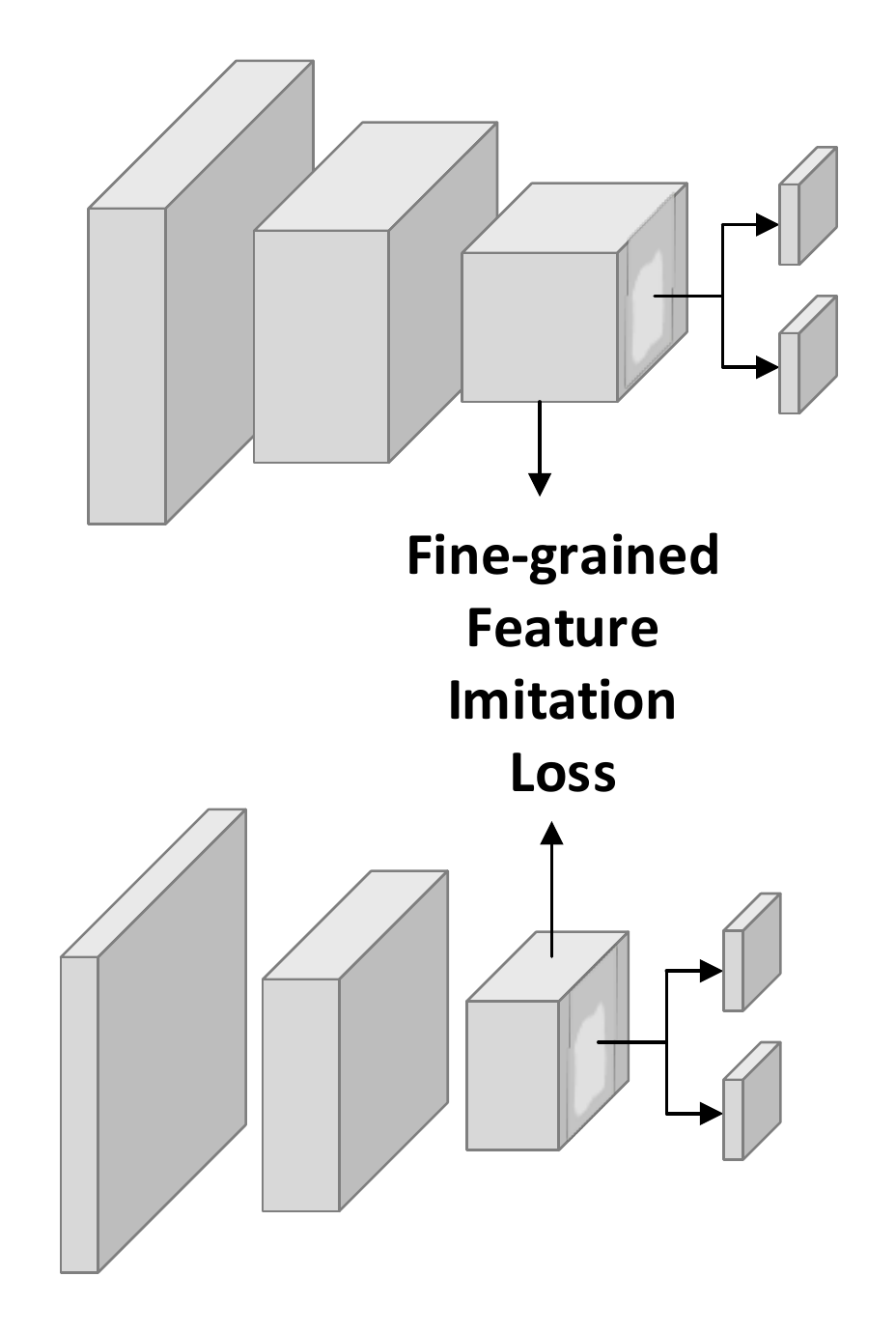}
        \end{minipage}%
    }
    ~
    \subfigure[\footnotesize \textbf{Ours}]{
        \label{fig:HGD_sample}
        \begin{minipage}[t]{0.43\linewidth}
            \centering
            \includegraphics[width=0.9\linewidth]{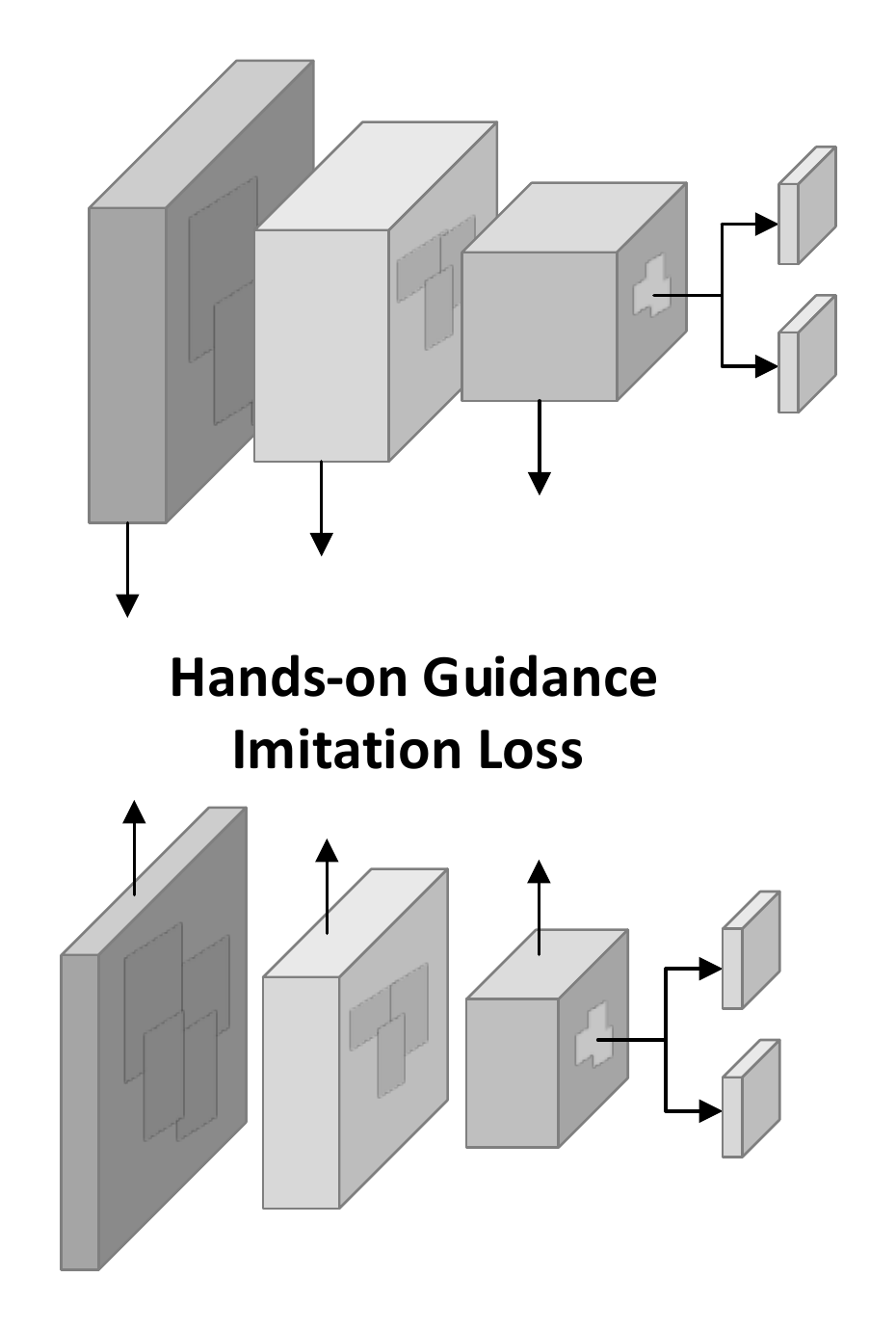}
        \end{minipage}
    }

    \vspace{-2mm}
    \caption{Framework comparison between different KD models and our HGD. And our method tries to do extreme feature imitation inspired by the evolution. }
    \label{fig:multi_pic}
    \vspace{-3 mm}

\end{figure}

\section{Method}
\label{sec:method}

In general, knowledge distillation is used as additional supervision besides the ground-truth. Previous KD works~\cite{wang2019distilling,chen2017learning,li2017mimicking} on detection mostly focus on high-level features or soft-label imitation, as dedicated in Figure \textcolor{red}{2(a,b,c)}. But we argue that the early-stage features equally make sense with distilling fine-grained location information.

 To validate the above argument, we conduct a preliminary experiment of different stages imitation on the SSD detector, as experimental details described in Section~\ref{sec:experiment}. For fairness, they have the same experimental conditions except for different stage choices. As shown in Table \textcolor{red}{\ref{exp:preliminary}}, imitating at an early stage yields even better results than later imitation, and it works best with all stage features. Guided by this observation, taking full advantage of the prior feature knowledge may have more potential for object detector distillation.

In the following sections, we introduce the details of designed \emph{Hands-on Guidance Distillation} framework.
We firstly introduce the feature matching strategy and then provide the details of imitation loss.
Finally, we describe how to integrate multiple imitation losses for joint optimization.

\subsection{Hands-on Guidance Distillation}

Our HGD is designed to distill the knowledge derived from the whole extracted features step by step.
 As applied to the conventional detector, an illustration of the design difference can be found in Figure \textcolor{red}{\ref{fig:multi_pic}}. Core to our approach is catalyzing the imitation supervision to the extreme, especially for feature-sensitive detection tasks.  It is equivalent to imposing more restrictions and can be regarded as a regularization method with teachers' help in a sense.
 On the whole, we start with a one-to-one correspondence between teacher network and student network based on semantic compatibility and then calculate the distance loss between them.
 And a natural question is how to deal with so many feature maps to ensure a particular part of the learning is purposefully emphasized.

\begin{table}[]
\caption{Preliminary VOC0712 experiments for direct imitation at different stages. Teacher model is the standard SSD~\cite{liu2016ssd} detector, and student model is the channel-half SSD.}
\label{exp:preliminary}
\small
\centering
\begin{tabular}{c|cccc}
\toprule
Stage & None  & Conv1$\sim$Conv4 & Conv5$\sim$Conv10 & All  \\
\midrule
      mAP(\%) &  65.6 & 70.2        &  70.0                 & 71.4 \\

\bottomrule
\end{tabular}

\vspace{-3mm}

\end{table}


\subsection{Learning}
\label{sub_sec:learning}
\subsubsection{Matching Method}

\begin{figure}[t]
\centering
\includegraphics[width=1.0\linewidth]{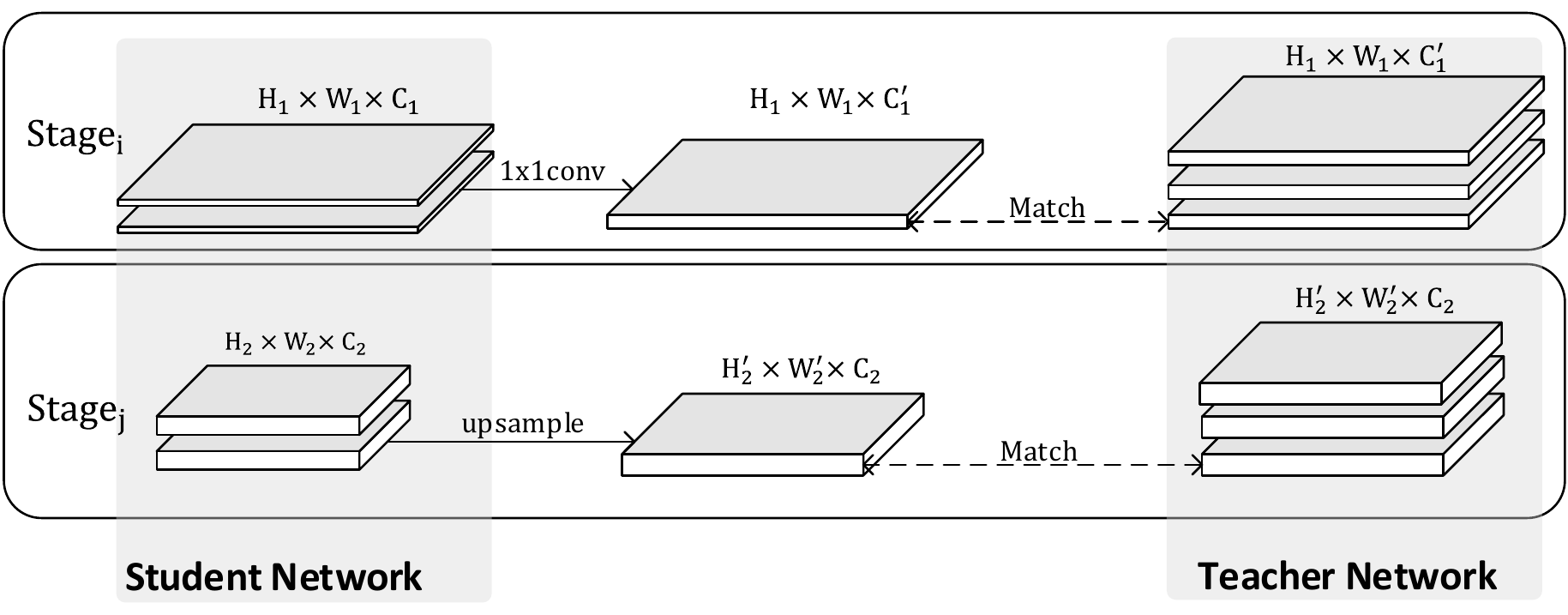}
\vspace{-9mm}
\caption{Illustration of matching method.
The last feature of each stage is selected and adapted before matching. Adaption operations include 1x1 conv, up-sample and down-sample.
}
\vspace{-4mm}
\label{fig:matching}
\end{figure}

First of all, we must set up the correspondence between student model and teacher model.
It's a key issue that using all features will inevitably lead to excessive memory consumption and slow convergence during training.
Moreover, the number of their feature maps may not match.
Therefore, we sample at each feature extraction stage for a one-to-one semantic correspondence in the stage dimension, as outlined in Figure \textcolor{red}{\ref{fig:matching}}.
Specifically, we divide all feature maps into different stages according to the spatial size.
And the last feature map of each stage is selected out for its richest semantics.
Then we pair these selected representatives separately. 
Besides, we add full convolution adaptation layers after the corresponding student features such that the channel number can be compatible with the teacher. 
The whole process can be formulated as:
$X'_{imit} =  adaption( select(X_{stage}))$  
where $X_{stage}$ and $X'_{imit}$ mean origin feature maps and compatible ones.
In terms of the particular case of incompatible input resolution, we use bilinear interpolation to do up-sample or down-sample adaption for unified resolution, or ignore some unmatched feature maps as alternatives. During deployment, these added adaption modules will be removed. 

\subsubsection{Loss Function}

After the correspondence and the adaption of features, we calculate the feature distance loss to ensure that the student model optimizes towards the teacher model with guidance. Generally, many researchers~\cite{wang2019distilling,zhu2019mask} use the $L2$ distance between student feature vector $V$ and teacher vector $Z$ in each coordinate as imitation loss $\mathcal{L} _{imit\_l2}$:
\begin{equation}
\mathcal{L} _{imit\_l2}(V,Z) = \left \|V - Z \right \|_{2}^{2}
\label{eq:L2}
\end{equation}
However, we argue that distilling in this way is too strict because it’s hard to force the student model to be the same as the teacher in every stage. Therefore, we adopt weaker supervision instead to avoid the excessive pursuit of consistency:
\begin{equation}
\mathcal{L} _{imit\_cosine}(V,Z) = 1 - cos(V,Z) = 1 - \frac{V\cdot Z}{\left \| V \right \|\left \| Z \right \| }
\label{eq:cosine}
\end{equation}
This cosine similarity loss $\mathcal{L} _{imit\_cosine}$ aims to grasp the vector direction of the teacher feature while strengthen guidance through more multi-stage imitation.
The experimental effect of replacing $L2$ with cosine loss can be seen in Table \textcolor{red}{\ref{exp:ablation-study}}.
Together with the features of all stages and all locations, the hands-on imitation loss $\mathcal{L} _{i}$ is to minimize:
\begin{equation}
\mathcal{L} _{i} =  \sum_{s=1}^{S}(\frac{1}{H_{s}W_{s}}\sum_{i=1}^{H_{s}}\sum_{j=1}^{W_{s}} \mathcal{L} _{imit}(V^s_{i,j},Z^s_{i,j}))
\label{eq:imitation}
\end{equation}
Where $S$ is the number of stages, $H_{s}$ and $W_{s}$ are the height and width of each stage feature map, $V^s_{i,j}$ and $Z^s_{i,j}$ denote the student and the teacher feature vector at location $(i,j)$ of stage $s$, respectively. Then we merge it with the common classification loss $\mathcal{L}_{c}$ and localization loss $\mathcal{L}_{l}$, which compose of a multi-task loss function as $\mathcal{L} = \mathcal{L}_{c} + {\lambda}_{1} \mathcal{L}_{l} + {\lambda}_{2} \mathcal{L}_{i}$ to jointly optimize parameters.
Adjust the weight ${\lambda}_{1}$ and ${\lambda}_{2}$ so that all loss items are at the same order of magnitude.

\subsubsection{Re-weighted Imitation}
\label{sec:re_weighted}
Even with sampling, it can be observed that the number of features to be imitated is still quite large, so that the student model may not concentrate on the emphasis. Hence we introduce a re-weighted hands-on imitation loss from both micro and macro perspectives in the sense that it makes HGD more targeted, and reformulate Eq.~\ref{eq:imitation} to elaborate as follows:
\begin{equation}
\mathcal{L} _{i} =  \sum_{s=1}^{S} u^{s} (\frac{1}{ \sum_{i,j}^{}v^{s}_{i,j}}\sum_{i=1}^{H_{s}}\sum_{j=1}^{W_{s}} v^{s}_{i,j} \mathcal{L} _{imit}(V^s_{i,j},Z^s_{i,j}))
\label{eq:re-weighted}
\end{equation}
 Where the added $u^{s}$ and $v^{s}_{i,j}$ indicate the macro stage weight of each stage $s$ and the micro spatial weight at each location $(i, j)$ of stage $s$, respectively.
 What’s more, each stage feature imitation loss is normalized by $\sum_{i,j}^{}v^{s}_{i,j}$ for the reason that different picture input may lead to entirely different spatial weight distribution. Another natural question is how to get these weights. Our intuition is to assess the importance of features from both overall and partial, and some strategies are considered for this purpose:
\\
\textbf{Attention.}
 Our initial idea was to use a learnable attention module to estimate their importance. However, it actually leads to abnormally sparse weights during training. Specifically, the weights are all close to 0 for the imitation losses that are difficult to optimize, and the weights of relatively simple ones become very large, resulting in final distillation basically not working even if adding some normalization constraints.
\\
\textbf{Focal weight.}
 Inspired by focal loss~\cite{lin2017focal}, our another attempt is to down-weight the well-imitated parts by adding an exponential term: $(1-p_{l_{i}}) ^{\gamma}l_{i}$
 where $p_{l_{i}}$ is the ratio of the loss value $l_{i}$ to total ones, $\gamma $ is a tunable parameter and we set $\gamma=2$.
\\
\textbf{Vector mean or variance.}
 Some network pruning works~\cite{han2015learning,han2015deep} maintain larger weights because they play a more critical role than smaller ones. Accordingly, we propose to calculate the mean statistics or variance statistics of each spatial vector $V$ for generating the spatial weighting score $v$:
 \begin{equation}
 v = sigmoid(statistics(V))
 \end{equation}
 and combine into the stage weight $u$:
  \begin{equation}
 u = mean(\left \{ v_{1},v_{2},..., v_{n} \right \})
  \end{equation}
 Note that using the statistics calculated on the original student features before the adaption has better experimental results.
\\
\textbf{GT-mask.}
 Considering paying more attention to the critical foreground areas can avoid irrelevant noise from the background, we can emphasize the object area in terms of the micro spatial weights. Specifically, we use $v=1$ for the foreground areas and $v=0$ for all others.
  And the areas inside the bounding box labels are regarded as foreground, while others outside as background. Besides, the macro stage weight $u$ can follow the mean or variance method mentioned above.


\section{Experiments}
\label{sec:experiment}

\subsection{Experimental Details}

We first introduce the model settings and datasets in this subsection, and then build extensive comparison experiments, including learning strategies, ablation study, related methods, different student models and different datasets. Moreover, we conduct some detailed analysis of the distillation process.
\\
\textbf{Models.}
 For fairness, all models are implemented on the PyTorch with  a uniform input resolution of $300\times300$ and follow the origin SSD settings~\cite{liu2016ssd}.
 In terms of distillation objects, we choose the origin VGG16 based SSD as teacher model. In contrast, the student model uses a weaker feature extractor, which has the same network architecture but only half channels left compared with the teacher model, called channel-half SSD. Other thin student models are also applicable, such as VGG11 and mobilenet~\cite{howard2017mobilenets} shown in Section \ref{sec:others_exps}.
\\
\textbf{Datasets.}
 We evaluate methods on commonly used PASCAL VOC0712 and MS COCO detection datasets, and all quantitative results are assessed in mean average precision (mAP) metric following convention. For distillation, the effect is reflected in the accuracy improvement of student models.
\\
\textbf{Training.}
 As prepared, the teacher model is trained in advance, and we fix all its parameters in the subsequent distillation. During training, we make student model learn by itself in the first few epochs for building on the detection foundation and then carry on distillation.
 In deployment, all except the original student model are removed.
 Note that teacher models are ImageNet pre-trained, while student models are not. And we slightly increase the number of iterations to guarantee sufficient training for introducing the distillation task.
%
\begin{table}
\setlength{\tabcolsep}{2.3mm}{
  \caption{ Comparison of different re-weighting choices.}
  \label{exp:design_choice}
  \centering
  \small
  \begin{tabular}{cccccc}
    \toprule
    \multicolumn{2}{c}{Re-weighting}  &Individual  &\multicolumn{3}{c}{Combination}\\

    \midrule
    \hline
    \multirow{4}*{Macro} &Attention   &67.1 &        &    &        \\
                         &Focal       &71.1 &\checkmark  &     &         \\
                         &Stage mean  & 71.8  & &\checkmark   &       \\
                         &Stage variance& 71.3  &    &       & \checkmark                    \\
    \hline
    \multirow{3}*{Micro} &Spatial mean       &71.5  &     &     &              \\
                         &Spatial variance   &71.7  &     &     &                        \\
                         &GT-mask            &72.2  &\checkmark &\checkmark      &\checkmark       \\
    \hline
     & &mAP(\%) &72.2 &72.3   &{\bf72.6}  \\
   \bottomrule
  \end{tabular}
}
\vspace{-3 mm}
\end{table}
%
\subsection{Imitation Effect}
In addition to the preliminary investigations of stages imitation in Table \textcolor{red}{\ref{exp:preliminary}}, where no re-weighting is used, a series of comparison experiments are conducted on VOC0712:
%
%
%
%
%
\begin{table*}[h]
\caption{Comparison with other SOTA detection distillation methods on VOC0712. The first two rows show the baseline trained with ground-truth supervision and the rest show the distillation effects.
The -half means the pruned channel-half VGG16.
}
\label{exp:sota}

\setlength{\tabcolsep}{0.45mm}{
\small
    \centering
    \begin{tabular}{c|cccccccccccccccccccc|c}
    \toprule
    {Methods} &aero &bike &bird &boat &bottle &bus &car &cat &chair &cow &table &dog &horse &mbike &person &plant &sheep &sofa &train &tv & {mAP(\%)} \\
    \midrule
    \hline
    SSD  teacher &79.6 &85.3 &75.0 &69.2 &50.4 &85.1 &86.7 &85.9 &60.1 &80.1 &75.3 &82.3 &86.1 &84.5 &79.7 &53.4 &74.4 &77.6 &86.4 &75.8    & 76.7                     \\
    SSD-half student   &70.5 &78.3 &54.3 &57.1 &27.6 &76.1 &82.5 &75.4 &46.6 &59.9 &68.4 &71.5 &78.8 &77.8 &74.7 &38.3 &61.8 &65.7 &78.4 &68.5  & 65.6   \\
    \hline

    Hint Learning\cite{romero2014fitnets}    &71.5 &80.0 &58.9 &59.0 &33.8 &79.7 &83.6 &78.3 &50.9 &66.6 &69.3 &74.4 &81.3 &80.5 &75.2 &43.0 &63.3 &66.9 &80.7 &66.6   & 68.2(+2.6)                                       \\

    Fine-Grained\cite{wang2019distilling}   &73.0 &79.9 &62.7 &59.6 &36.3 &80.9 &82.8 &81.2 &52.3 &70.4 &72.4 &77.4 &83.2 &80.2 &76.1 &44.0 &65.7 &73.1 &82.6 &68.8    & 70.1(+4.5)                                         \\
    Mask Guided\cite{zhu2019mask}  &71.2 &81.9 &63.5 &59.9 &37.1 &78.6 &82.6 &80.3 &53.4 &72.0 &74.6 &76.7 &83.4 &79.6 &75.9 &44.6 &65.2 &72.6 &82.6 &68.7   & 70.2(+4.6)                                          \\
    Our HGD    &73.2 &82.2 &67.6 &61.9 &39.3 &83.2 &82.9 &83.5 &55.4 &73.9 &73.0 &79.9 &84.9 &82.1 &77.0 &48.0 &71.3 &76.7 &85.1 &70.5               & \textbf{72.6(+7.0)}\\
    \bottomrule
    \end{tabular}
}
\vspace{-4 mm}
\end{table*}
%
%
%
\\
\textbf{Re-weighting strategies.}
 As shown in Table \textcolor{red}{\ref{exp:design_choice}}, we explore the impact of proposed re-weighting strategies mentioned in Section \ref{sec:re_weighted}. The first column represents different strategies from macro and micro perspectives, the second column reports their individual effect in HGD, and the rest columns report some combinations of them for extreme effect.
 Note that we choose SSD as the teacher model and channel-half SSD as the student model for all the experiments in this subsection.
 The results show that the combination of concentrating on both the critical foreground areas and the stages with large variance is superior to other strategies.
 %
%
%
%
\begin{table}
\setlength{\tabcolsep}{2mm}{
  \caption{Ablation study of HGD performance on VOC. *:The origin is imitating on detection head features with L2 metric. }
  \label{exp:ablation-study}
  \centering
  \small
  \begin{tabular}{ccccccc}
    \toprule
    &\multicolumn{6}{c}{From ordinary distillation to HGD}    \\
    \midrule
    \hline
    Hands-on&      &\checkmark  &\checkmark   &\checkmark   &\   &\checkmark      \\
    Cosine metric&      &       &\checkmark   &\checkmark   &\checkmark    &\checkmark       \\
    Feature selection &    &        &             &\checkmark   &\checkmark &\checkmark     \\
    Re-weighted      &        &     &               &   &\checkmark              &\checkmark         \\
    \hline
    {mAP(\%)} &68.1* &68.3   &69.0  &71.4 &70.8  &{\bf72.6} \\
   \bottomrule
  \end{tabular}
}
\vspace{-4 mm}
\end{table}
\\
\textbf{Comparison with state-of-the-art.}
 We present a comparison with the SOTA that mostly focus on high-level features. As shown in Table \textcolor{red}{\ref{exp:sota}}, HGD surpasses others by more than 2.4\% and recovers 63\% precision degradation from cutting channels, which demonstrates our method can effectively distill the teacher detector’s knowledge into the halved student.
\\
\textbf{Ablation study.}
  To measure the advantage gained, we progressively increase modules from the origin in Table \textcolor{red}{\ref{exp:ablation-study}}.
  Note that the hands-on approach imitates all features instead if the feature selection is not active.
  The results show that each strategy is crucial to the final. For example, the hands-on approach only brings a little improvement when used alone, but the final combination will drop a lot without it (last two columns). These well-designed modules are appropriate to complement each other, making our method more effective.

\subsection{More Challenging Experiments}
\label{sec:others_exps}

 To verify the generalization ability of the HGD, we conduct further verification on more difficult distillation scenes.
\\
\textbf{Different teacher-student patterns.}
 In addition to channel-half VGG16 backbone, VGG11 and mobilenet backbones are also compatible. As shown in Table \textcolor{red}{\ref{exp:TS_patterns}}, VGG11 gains 2.2\% mAP, and mobilenet gains 3.1\% with HGD. 
 Different mAP gains result from the different fitting capabilities of original student networks, and
 the final effect has an upper limit.
%
%
%
\begin{table}
\setlength{\tabcolsep}{0.85mm}{    
  \caption{Distill different student models with VGG16-SSD. *:Single batch test on NVIDIA Titan Xp, NMS not included.}
  \label{exp:TS_patterns}
  \centering
  \small
  \begin{tabular}{cccccc}
    \toprule
    {Backbone} & MParams &GFLOPs &FPS*  &Origin mAP & After HGD  \\
    \midrule
    \hline
    VGG16 &26.29 &31.44 &98 &76.7 & -\\

    VGG16-half  &7.41 &8.23 &121 &65.6 &72.6(+7.0)\\
    VGG11 &20.79  &17.20 &128 &72.7 &74.9(+2.2)\\
    MobileNet &4.4 &1.19 &72  &62.5 &65.6(+3.1)\\
    \bottomrule
  \end{tabular}
}
\vspace{-4 mm}
\end{table}
%
%
%
\\
\textbf{Results on MS COCO.}
 We further test on more challenging MS COCO dataset, which contains much smaller and hard distinguishable  objects. 
 As dedicated in Table \textcolor{red}{\ref{exp:COCO}}, the SSD-half student model gains 1.9 absolute mAP with our HGD, while the contrast method is basically ineffective, verifying the robustness of handling more complicated examples.

\begin{table}
\setlength{\tabcolsep}{0.8mm}{    
  \caption{Results of distillation on MS COCO.}
  \label{exp:COCO}
  \centering
  \small
  \begin{tabular}{cccccc}
    \toprule

    \multirow{2}*{Method} & \multirow{2}*{MParams} &\multirow{2}*{GFLOPs}  &\multicolumn{3}{c}{AP (\%), IOU} \\
    \cmidrule{4-6}
    &  &  &0.5 &0.75 &0.5:0.95 \\
    \midrule
    \hline
    SSD teacher  &  34.31 &34.42    &39.5 &24.1 &23.4\\
    SSD-half student &11.42 & 9.72  &30.0 &16.8 &16.8\\
    \hline
    student+Mask Guided\cite{zhu2019mask}  & 11.42 & 9.72 &30.2  &17.0  &17.1 \\
    student+HGD & 11.42 & 9.72 & 32.8  &18.8  &18.7   \\
    \bottomrule
  \end{tabular}
}
\vspace{-2 mm}
\end{table}

\subsection{Analysis}

\begin{figure}[]
\centering
\includegraphics[width=1.0\linewidth]{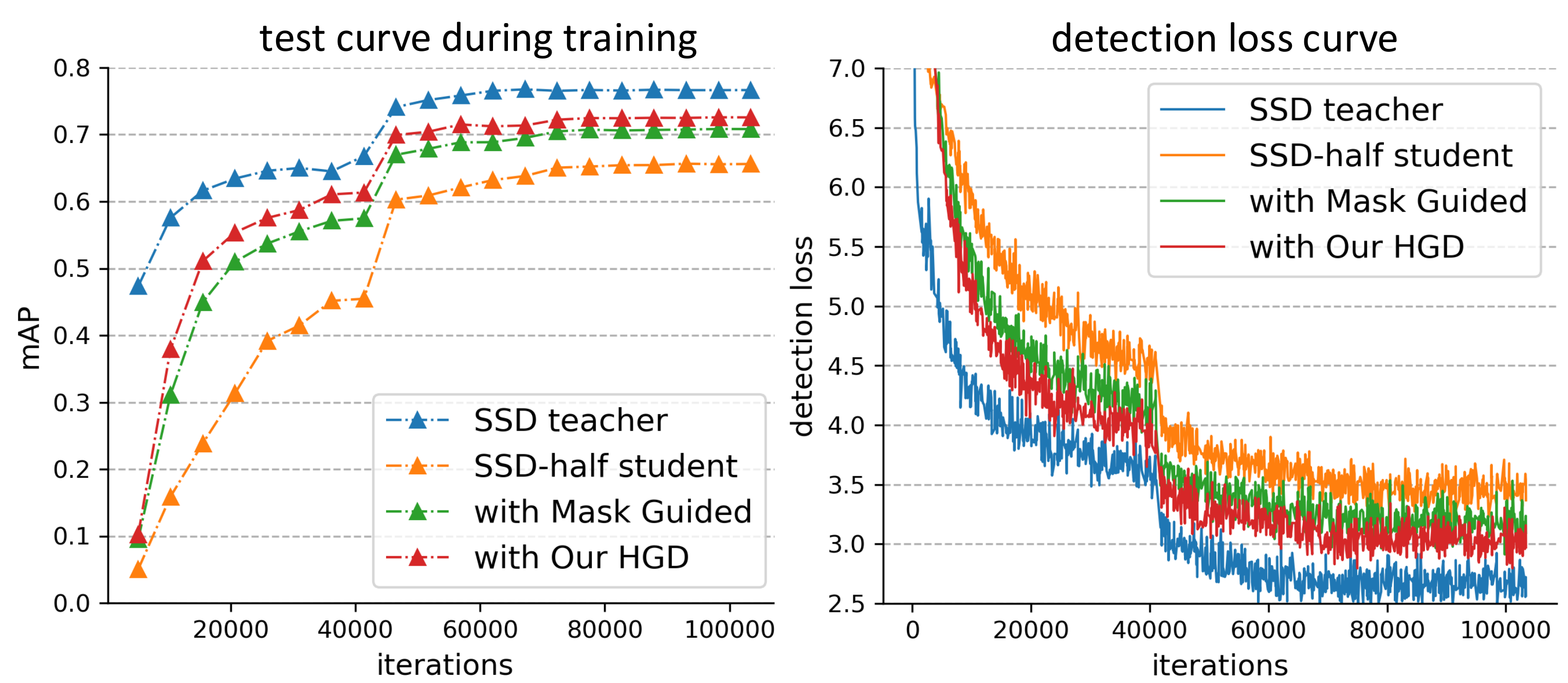}
\vspace{-8mm}
\caption{Illustration of training before and after distillation. The turning points results from the step decay LR strategy.}
\label{fig:mAP_loss_Curves}
\end{figure}

\begin{figure}[]
\centering
\includegraphics[width=1.0\linewidth]{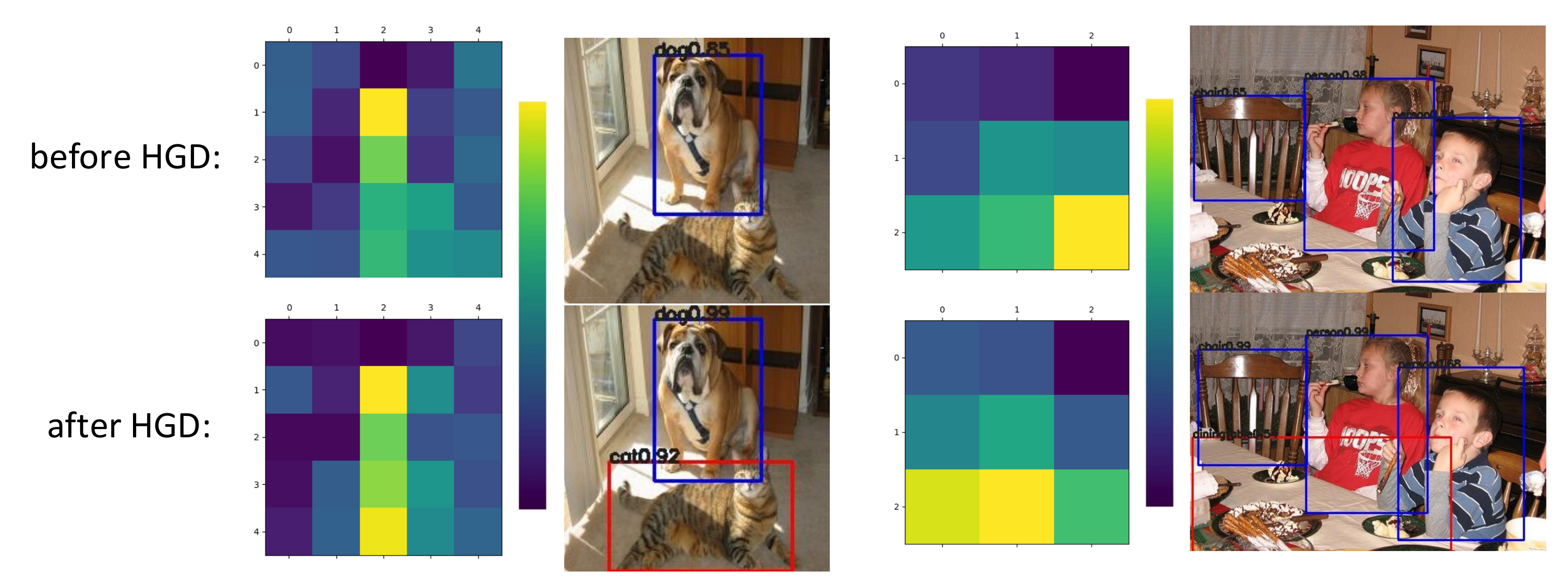}
\vspace{-5mm}
\caption{Visualization of Conv9\_2 or Conv10\_2 feature map. Some missing objects are recalled for more accurate features.}
\label{fig:Vis}
\vspace{-2 mm}
\end{figure}

To better understand the process of knowledge distillation, we show the precision and the partial loss items in Figure \textcolor{red}{\ref{fig:mAP_loss_Curves}}.
Same as mAP, the detection loss, including only the localization loss $\mathcal{L}_{l}$ and the classification loss $\mathcal{L}_{c}$, can quantitatively assess detection capabilities ($\mathcal{L}_{l}$ and $\mathcal{L}_{c}$ use the same loss function while $\mathcal{L}_{i}$ is different in comparison).
Compared with the baseline and previous methods, the optimization curves of detection loss and precision converge faster with HGD,  and the final mAP is higher. These demonstrate that HGD can effectively accelerate convergence and promote more knowledge absorption, further achieving the purpose of distillation.

 Moreover, we choose some intermediate features in SSD-half student for visualization in Figure \textcolor{red}{\ref{fig:Vis}}, where the color of each grid represents the activation intensity of the corresponding location. With HGD, the features of critical locations are enhanced, which naturally leads to better detection results.

\section{Conclusion}
In this paper, we propose a more comprehensive detector distillation method, referred as HGD, to distill the knowledge step by step instead of the higher-level feature tendency. With it, small detectors can fully imitate the feature hints, and the effects become much closer to the cumbersome teacher’s, which is significant for lightweight deployment. Extensive experiments on different models and datasets demonstrate the effectiveness of our method.
It is worth noting that our method only relies on feature maps, leading to more robust compatibility and development potential.

{\small
\bibliographystyle{IEEEbib}
\bibliography{qyy_icme2021}
}

\end{document}